# Strongly-Typed Agents are Guaranteed to Interact Safely

David Balduzzi [1]

## Abstract

As artificial agents proliferate, it is becoming increasingly important to ensure that their interactions with one another are well-behaved. In this paper, we formalize a common-sense notion of when algorithms are well-behaved: an algorithm is *safe* if it *does no harm*. Motivated by recent progress in deep learning, we focus on the specific case where agents update their actions according to *gradient descent*. The paper shows that that gradient descent converges to a Nash equilibrium in safe games. The main contribution is to define strongly-typed agents and show they are guaranteed to interact safely, thereby providing sufficient conditions to guarantee safe interactions. A series of examples show that strong-typing generalizes certain key features of convexity, is closely related to blind source separation, and introduces a new perspective on classical multilinear games based on tensor decomposition.

## 1. Introduction

*"First, do no harm"*

Recent years have seen rapid progress on core problems in artificial intelligence such as object and voice recognition (Hinton & et al, 2012; Krizhevsky et al., 2012), playing video and board games (Mnih et al., 2015; Silver et al., 2016), and driving autonomous vehicles (Zhang et al., 2016). As artificial agents proliferate, it is increasingly important to ensure their interactions with one another, with humans, and with their environment are *safe*.

Concretely, the number of neural networks being trained and used is growing rapidly. There are enormous and increasing economies of scale that can likely be derived from treating them as populations – rather than as isolated algorithms. How to ensure interacting neural networks cooperate effectively? When can weights trained on one problem be adapted to another without adverse effects? The problems fall under *mechanism design*, a branch of game theory (Nisan et al., 2007). However, neural nets differ from humans in that they optimize clear objectives using *gradient descent*. The setting is thus more structured than traditional mechanism design.

**Safety.** The first contribution of the paper is formalize safety as a criterion on how agents interact. We propose a basic notion of safety based on the common-sense principle that agents should do no harm to one another. More formally, each agent optimizes an objective whose value depends on the actions of the agent and the actions of the rest of the population. A game is safe if the actions chosen by each agent do no (infinitesimal) harm to any other agent, where harm is measured as increased loss.

The key simplifying assumption in the paper is to **take gradient descent as a computational primitive** (Balduzzi, 2016). Questions about mechanism design are sharpened under the assumption that agents use gradient descent. The assumption holds broadly since the key driver of progress in artificial intelligence is deep learning, which uses gradient descent to optimize complicated objective functions composed from simple differentiable modules (LeCun et al., 2015).

A weakness of the approach is that it conceives safety more narrowly than, for example, Amodei et al. (2016) which is concerned with societal risks arising from artificial intelligence. We argue that a necessary foundational component of the broader AI-safety project is to clarify exactly what safety entails when the objectives of the agents and the algorithms they employ are precisely specified.

**Strongly-typed games.** The second contribution is to introduce type systems suited to multi-agent optimization problems (that is, games). We build on the typed linear algebra introduced in Balduzzi & Ghifary (2016). The nomenclature is motivated by an analogy with types in the theory of programming. Type systems are used to prevent untrapped errors (errors that go unnoticed and cause arbitrary behavior later on) when running a program (Cardelli, 1997). A program is safe if it does not cause untrapped errors. Type systems can enforce safety by statically rejecting all programs that are potentially unsafe.

[1] Victoria University of Wellington, New Zealand. Correspondence to: David Balduzzi <dbalduzzi@gmail.com>.

The idea underlying types in programming is that "like should interact with like". Typed linear algebra, definition 1, formalizes "like interacts with like" in the simplest possible way – by fixing an orthogonal basis. Section 2 introduces a wider class of games than in the literature and defines safety. Theorem 1 shows that gradient descent converges to a Nash equilibrium in safe games. Section 3 extracts the key ingredients required for safe gradients from two warmup examples. The ingredients are *simultaneous diagonalization*, i.e. the existence of a shared latent orthogonal basis, and *monotonic covariation*, i.e. that the derivatives of the objectives have the same sign in the latent coordinate system. The main result, theorem 2, is that *strongly-typed games* are guaranteed to be safe.

**Implications.** Safety and strong-typing generalize key properties of convexity. Convexity is of course the gold standard for well-behaved gradients. We uncover latent types and demonstrate safety of Newton's method, natural gradient and mirror descent; see sections 3.2, A2 and A3.

The main theme of sections 4 and 5 is *disentangling latent factors*. We show that strong-typing in quadratic games is closely related to blind source separation. Section 5 analyzes classical $N$-player games. The analysis yields a new perspective on classical games based on tensor-SVD that is closely related to independent component analysis.

Sections 6 and A6 switch to neural networks and analyze two biologically plausible variants of backpropagation (Balduzzi et al., 2015; Lillicrap et al., 2016). We show that the main results of the papers are to prove the respective algorithms are safe.

**Scope and related work.** This paper lays the foundations of safety in gradient-based optimization. Applications are deferred to future work.

The literature on safety is mostly focused on problems arising in reinforcement learning, for example ensuring agents avoid dangerous outcomes (Turchetta et al., 2016; Amodei et al., 2016; Berkenkamp et al., 2016). Gradients are typically not available in reinforcement learning problems. We study interactions between algorithms with clearly defined objectives that utilize gradient-based optimization, which gives a more technical perspective.

The idea of a population of neural networks solving multiple related tasks is developed in Fernando et al. (2017), which uses genetic algorithms to adapt components to new tasks. However, they repeatedly reinitialize components to undo the damage done by the genetic algorithm. Our work is intended, ultimately, to help design algorithms that detect and avoid damaging updates. A recent survey paper argues the brain optimizes a family of complementary loss functions (Marblestone et al., 2016) without considering how the complementarity of the loss functions could be checked or enforced.

The idea of investigating game-theoretic and mechanism design questions specific to certain classes of algorithms is introduced in Rakhlin & Sridharan (2013); Syrgkanis et al. (2015). The papers consider how convergence in games can be accelerated if the players use variants of mirror descent.

**Terminology.** If $\alpha \geq 0$ then $\alpha$ is *positive*; if $\alpha > 0$ then it is *strictly* positive. A (not necessarily square) matrix $\mathbf{D}$ is diagonal if $d_{ij} = 0$ for all $i \neq j$ and similarly for tensors. Vectors are columns. The inner product is $\langle \mathbf{v}, \mathbf{w} \rangle = \mathbf{v}^\intercal \mathbf{w}$.

## 2. Safety

### 2.1. Types and orthogonal projections

Let us recall some basic facts about orthogonal projections. Let $(V, \langle \bullet, \bullet \rangle)$ be a vector space equipped with an inner product. An **orthogonal projection** is a linear transform $\boldsymbol{\pi} : V \to V$ that is

O1. idempotent, $\boldsymbol{\pi}^2 = \boldsymbol{\pi}$, and

O2. self-adjoint, $\langle \boldsymbol{\pi} \mathbf{v}, \mathbf{v}' \rangle = \langle \mathbf{v}, \boldsymbol{\pi} \mathbf{v}' \rangle$ for any $\mathbf{v}, \mathbf{v}' \in V$.

**Lemma 1.** *Let $\mathbf{P}$ denote an $(n \times k)$-matrix with orthogonal columns $\mathbf{p}_1, \ldots, \mathbf{p}_k$. Then the $(n \times n)$-matrix $\mathbf{PP}^\intercal = \sum_{i=1}^{k} \mathbf{p}_i \langle \mathbf{p}_i, \bullet \rangle = \sum_{i=1}^{k} \mathbf{p}_i \mathbf{p}_i^\intercal$ is an (orthogonal) projection.*

**Lemma 2.** *If two orthogonal projections $\boldsymbol{\pi}$ and $\boldsymbol{\tau}$ commute then their product is an orthogonal projection.*

*Proof.* Let $\mathbf{q} := \boldsymbol{\pi}\boldsymbol{\tau}$. If $\boldsymbol{\pi}\boldsymbol{\tau} = \boldsymbol{\tau}\boldsymbol{\pi}$ then

$$\mathbf{q}^2 = \boldsymbol{\pi}\boldsymbol{\tau} \cdot \boldsymbol{\pi}\boldsymbol{\tau} = \boldsymbol{\pi}\boldsymbol{\pi} \cdot \boldsymbol{\tau}\boldsymbol{\tau} = \boldsymbol{\pi}\boldsymbol{\tau} = \mathbf{q}.$$

Checking self-adjointness is an exercise. □

**Definition 1.** *A **type** $\mathcal{T}_V = \big(V, \langle \bullet, \bullet \rangle, \{\boldsymbol{\pi}_r\}_{r=1}^{R}\big)$ is a $D$-dimensional vector space with an inner product and orthogonal projections $\boldsymbol{\pi}_r : V \to V$ such that $\boldsymbol{\pi}_r \boldsymbol{\pi}_s = 0$ for $r \neq s$ and $\sum_{r=1}^{R} \boldsymbol{\pi}_r = \mathbf{I}_V$ is the identity. Type $\mathcal{T}_V$ has **dimension** $D$ and **rank** $R$.*

A full rank type, $D = R$, is equivalent to a vector space equipped with an orthogonal basis. Lower rank types are less rigid, and can be thought of as vector spaces equipped with generalized orthogonal coordinates.

### 2.2. Safe games

**Definition 2.** *A **game** consists of a type $\mathcal{T}_V$, feasible set $\mathcal{H} \subset V$, players $[N] := \{1, \ldots, N\}$, losses $\ell_n : \mathcal{H} \to \mathbb{R}$, and an assignment $\rho : [N] \to [R]$ of players to projections.*

The type structure and assignments specify the coordinates controlled by each player. On round $t$, player $n$ chooses $\boldsymbol{\xi}_n^t \in V$ and updates the joint action via

$$\mathbf{w}^{t+1} = \mathbf{w}^t - \boldsymbol{\pi}_{\rho(n)}(\boldsymbol{\xi}_n^t) \quad \text{where} \quad \mathbf{w}^t, \mathbf{w}^{t+1} \in \mathcal{H}.$$

Updates leaving the feasible set can be mapped back into it, see section A1. The projection $\boldsymbol{\pi}_{\rho(n)}$ specifies the coordinates of the joint-action vector that player $n$ can modify.

*Example* 1. In a **block game** actions $\mathbf{w} \in V = \prod_{n=1}^N \mathbb{R}^{D_n}$ decompose as $\mathbf{w} = (\mathbf{w}_1, \ldots, \mathbf{w}_N)$ where the $n^{\text{th}}$ player can modify the coordinates in $\mathbf{w}_n$. The orthogonal projections $\boldsymbol{\pi}_n(\mathbf{w}) = (\mathbf{0}, \ldots, \mathbf{w}_n, \ldots, \mathbf{0})$ form a rank-$N$ type with $\rho(n) = n$ for all $n \in [N]$.

*Example* 2. In an **open game** the type has $\text{rank}(\mathcal{T}_V) = 1$ so the single projection is the identity and $\rho(n) = 1$ for all $n$. Every player can modify all the coordinates.

Block games coincide with the standard definition of a game in the literature. Open games arise below when considering Newton's method, natural gradients, mirror descent and neural networks.

The goal of each player is to minimize its loss. Safety is the condition that no player's updates harm any other player.

**Definition 3.** *It is **safe** for player $m$ to choose $\boldsymbol{\xi}_m^t \in V$ if it does no infinitesimal harm to any player*

$$\langle \boldsymbol{\pi}_{\rho(m)}(\boldsymbol{\xi}_m^t), \nabla \ell_n(\mathbf{w}^t) \rangle \geq 0 \text{ for all } n \in [N].$$

*A **game is safe** if it is safe for players to use gradient descent: i.e. if choosing $\boldsymbol{\xi}_m^t := \nabla \ell_m(\mathbf{w}^t)$ is safe for all $m$.*

It is worth getting a degenerate case out of the way. A block game is decomposable if player $m$'s loss only depends on the actions it controls. Intuitively, a decomposable game is $N$ independent optimization problems. More formally:

**Lemma 3.** *A block game is **decomposable** if $\ell_m(\mathbf{w}) = \ell_m(\boldsymbol{\pi}_m \mathbf{w})$ for all $\mathbf{w}$ and $m$. Decomposable games are safe.*

*Proof.* Since $\boldsymbol{\pi}_m$ is self-adjoint, we have that $\langle \boldsymbol{\pi}_m \boldsymbol{\xi}, \boldsymbol{\eta} \rangle = \langle \boldsymbol{\pi}_m \boldsymbol{\xi}, \boldsymbol{\pi}_m \boldsymbol{\eta} \rangle$. Decomposability implies $\boldsymbol{\pi}_m(\nabla \ell_n) = 0$ when $m \neq n$, so

$$\langle \boldsymbol{\pi}_m(\nabla \ell_m), \boldsymbol{\pi}_m(\nabla \ell_n) \rangle = \begin{cases} \|\boldsymbol{\pi}_m(\nabla \ell_m)\|_2^2 & \text{if } m = n \\ 0 & \text{else} \end{cases}$$

which is always positive. $\square$

### 2.3. Convergence

A block game is **convex** if the feasible set $\mathcal{H}$ is compact and convex and the losses $\ell_n : \mathcal{H} \to \mathbb{R}$ are convex in the coordinates controlled by the respective players. Nash equilibria are guaranteed to exist in convex block games (Nash, 1950). However, finding them is often intractable (Daskalakis et al., 2009). We show gradient descent converges to a Nash equilibrium in safe convex games.

**Theorem 1.** *Gradient descent converges to a Nash equilibrium in safe convex games with smooth losses.*

*Proof.* Introduce potential function $\Phi(\mathbf{w}) = \sum_{n=1}^N \alpha_n \cdot \ell_n(\mathbf{w})$ where $\alpha_n > 0$ are strictly positive. Then

$$\langle \boldsymbol{\pi}_m(\nabla \Phi), \nabla \ell_m \rangle = \sum_{n=1}^N \alpha_n \langle \boldsymbol{\pi}_m(\nabla \ell_n), \nabla \ell_m \rangle \quad (1)$$
$$\geq \alpha_m \cdot \|\boldsymbol{\pi}_m(\nabla \ell_m)\|_2^2 \geq 0$$

since safety implies the cross-terms are nonnegative. The players' updates therefore converge to either a critical point of $\Phi$ or to the boundary of the feasible set. Suppose gradient descent converges to the interior of $\mathcal{H}$. Eq (1) implies that if $\nabla \Phi = 0$ then $\boldsymbol{\pi}_m(\nabla \ell_m) = 0$ for all $m$. By convexity of the losses, the critical point is a minimizer of each loss with respect to that player's actions, implying it is a Nash equilibrium. A similar argument holds if gradient descent converges to the boundary, see section A1. $\square$

*Example* 3 (convergence in a safe constrained game). Consider a two-player block game with $\ell_1(x, y) = x + 2y$ and $\ell_2(x, y) = 2x + y$ where player-1 controls $x$ and player-2 controls $y$. Introduce feasible set $\mathcal{H} = \{(x, y) \in \mathbb{R}^2 : x^2 + y^2 \leq 1\}$. The game is convex and safe. The set of Nash equilibria is the bottom-left quadrant of the boundary $\{(x, y) \in \mathcal{H} : x, y \leq 0 \text{ and } x^2 + y^2 = 1\}$. Gradient descent with positive combinations of $\boldsymbol{\pi}_1(\nabla \ell_1) = \frac{\partial}{\partial x}$ and $\boldsymbol{\pi}_2(\nabla \ell_2) = \frac{\partial}{\partial y}$ always converges to a Nash equilibrium.

A simple game that does *not* converge is the following zero-sum game, which is related to generative adversarial networks (Goodfellow, 2017).

*Example* 4 (convergence requires positivity). Consider the two-player block game $\ell_1(x, y) = xy$ and $\ell_2(x, y) = -xy$ where player-1 controls $x$ and player-2 controls $y$. The Nash equilibrium is the origin $(x, y) = (0, 0)$. However, gradient descent does not converge. Observe that $\nabla \ell_1 = y \frac{\partial}{\partial x} + x \frac{\partial}{\partial y}$ and $\nabla \ell_2 = -y \frac{\partial}{\partial x} - x \frac{\partial}{\partial y}$ so $\boldsymbol{\pi}_1(\nabla \ell_1) = y \frac{\partial}{\partial x}$ and $\boldsymbol{\pi}_2(\nabla \ell_2) = -x \frac{\partial}{\partial y}$. The flow $\boldsymbol{\pi}_1(\nabla \ell_1) + \boldsymbol{\pi}_2(\nabla \ell_2)$ rotates around the origin. No positive combination of $\boldsymbol{\pi}_1(\nabla \ell_1)$ and $\boldsymbol{\pi}_2(\nabla \ell_2)$ converges to the origin.

## 3. Strongly-Typed Games

Strong-typing is based two key ideas: diagonalization and positivity. Diagonalization is an important tool in applied mathematics. The Fourier transform simultaneously diagonalizes differentiation and convolution:

$$\mathcal{F}\left(\frac{df}{dx}\right) = 2\pi i \omega \mathcal{F}(f) \quad \text{and} \quad \mathcal{F}(f * g) = \mathcal{F}(f) \cdot \mathcal{F}(g)$$

The SVD diagonalizes any matrix: $\mathbf{Q}^\intercal \mathbf{MP} = \mathbf{D}$. Finally, the Legendre transform $f^*(\boldsymbol{\eta}) = \max_{\boldsymbol{\theta}}\{\boldsymbol{\eta}^\intercal \boldsymbol{\theta} - f(\boldsymbol{\theta})\}$ diagonalizes the infimal convolution

$$(f \square g)^* = f^* + g^* \text{ for } (f \square g)(\boldsymbol{\theta}) = \min_{\boldsymbol{\vartheta}}\{f(\boldsymbol{\vartheta}) + g(\boldsymbol{\theta} - \boldsymbol{\vartheta})\}.$$

Diagonalization finds a latent orthogonal basis that is more mathematically amenable than the naturally occurring coordinate system. Strong-typing is based on an extension of diagonalization to nonlinear functions. Before diving in, we recall the basics of simultaneous diagonalization.

**Symmetric matrices.** Any symmetric matrix $\mathbf{A}$ factorizes as $\mathbf{A} = \mathbf{P}^\intercal \mathbf{DP}$ where $\mathbf{P}$ is orthogonal and $\mathbf{D}$ is diagonal. A collection $\mathbf{A}_1, \ldots, \mathbf{A}_N$ of symmetric matrices is simultaneously diagonalizable iff the matrices commute, in which case $\mathbf{A}_i = \mathbf{P}^\intercal \mathbf{D}_i \mathbf{P}$ where $\mathbf{D}_i$ is diagonal and $\mathbf{P}$ determines a common latent coordinate system (or type).

**Arbitrary matrices.** The diagonalization of an $(m \times n)$-matrix $\mathbf{A}$ is $\mathbf{A} = \mathbf{PDQ}^\intercal$ where $\mathbf{P}$ and $\mathbf{Q}$ are orthogonal $(m \times m)$ and $(n \times n)$ matrices and $\mathbf{D}$ is positive diagonal. A collection of matrices is simultaneously diagonalizable if $\mathbf{A}_i = \mathbf{PD}_i \mathbf{Q}^\intercal$ for all $i$. A necessary condition for simultaneous diagonalizability is that

$$\mathbf{A}_i^\intercal \mathbf{A}_j \text{ and } \mathbf{A}_i \mathbf{A}_j^\intercal \text{ are symmetric for all } i, j. \quad (2)$$

Next, we work through two examples where diagonalization and a positivity condition imply safety.

### 3.1. Warmup: When are two-player games safe?

To orient the reader, we consider a minimal example which illustrates most of the main ideas of the paper: two-player bilinear games (von Neumann & Morgenstern, 1944). Consider a two-player block game with loss functions

$$\ell_1(\mathbf{v}, \mathbf{w}) = \mathbf{v}^\intercal \mathbf{Aw} \quad \text{and} \quad \ell_2(\mathbf{v}, \mathbf{w}) = \mathbf{v}^\intercal \mathbf{Bw}$$

and projections $\boldsymbol{\pi}_{1/2}(\mathbf{v}, \mathbf{w}) = (\mathbf{v}, \mathbf{0})$ and $(\mathbf{0}, \mathbf{w})$. The gradients are $\nabla \ell_1 = \sum_{ij}(w_j A_{ij} \frac{\partial}{\partial v_i} + v_i A_{ij} \frac{\partial}{\partial w_j}) = (\mathbf{w}^\intercal \mathbf{A}^\intercal, \mathbf{v}^\intercal \mathbf{A})$ and $\nabla \ell_2 = (\mathbf{w}^\intercal \mathbf{B}^\intercal, \mathbf{v}^\intercal \mathbf{B})$. The game is safe if

$$\langle \boldsymbol{\pi}_1(\nabla \ell_1), \nabla \ell_2 \rangle = \mathbf{w}^\intercal \mathbf{A}^\intercal \mathbf{Bw} \geq 0 \quad \text{and}$$
$$\langle \nabla \ell_1, \boldsymbol{\pi}_2(\nabla \ell_2) \rangle = \mathbf{v}^\intercal \mathbf{BA}^\intercal \mathbf{v} \geq 0 \quad \text{for all } \mathbf{v} \text{ and } \mathbf{w}.$$

Safety requires that $\mathbf{A}^\intercal \mathbf{B}$ and $\mathbf{BA}^\intercal$ are positive semidefinite. Any square matrix decomposes into symmetric and antisymmetric components $\mathbf{M} = \mathbf{M}^s + \mathbf{M}^a = \frac{1}{2}(\mathbf{M} + \mathbf{M}^\intercal) + \frac{1}{2}(\mathbf{M} - \mathbf{M}^\intercal)$ where $\mathbf{w}^\intercal \mathbf{M}^a \mathbf{w} = 0$ for all $\mathbf{w}$. Thus, a square matrix is positive semidefinite iff its symmetric component is positive semidefinite.

We therefore restrict to when $\mathbf{A}^\intercal \mathbf{B}$ and $\mathbf{BA}^\intercal$ are symmetric. Recalling (2), we further suppose that $\mathbf{A}$ and $\mathbf{B}$ are simultaneously diagonalizable and obtain:

**Lemma 4.** *A two-player game is safe if $\mathbf{A} = \mathbf{PDQ}^\intercal$ and $\mathbf{B} = \mathbf{PEQ}^\intercal$ where $\mathbf{P}$ and $\mathbf{Q}$ are orthogonal matrices, $\mathbf{D}$ and $\mathbf{E}$ are diagonal, and $\mathbf{DE} \geq 0$.*

*Proof.* The assumptions imply that

$$\langle \boldsymbol{\pi}_1 \nabla \ell_1, \nabla \ell_2 \rangle = \mathbf{w}^\intercal \mathbf{A}^\intercal \mathbf{Bw} = \mathbf{w}^\intercal \mathbf{Q}(\mathbf{D}^\intercal \mathbf{E})\mathbf{Q}^\intercal \mathbf{w} \geq 0$$

and $\langle \nabla \ell_1, \boldsymbol{\pi}_2 \nabla \ell_2 \rangle = \mathbf{w}^\intercal \mathbf{P}(\mathbf{E}^\intercal \mathbf{D})\mathbf{P}^\intercal \mathbf{w} \geq 0$. □

### 3.2. Warmup: When is Newton's method safe?

It was observed in Dauphin et al. (2014) that applying Newton's method to neural networks is problematic because it is attracted to saddle points and can *increase* the loss on nonconvex problems. We reformulate their observation in the language of safety.

Consider a single player open game with twice differentiable loss $\ell : V \to \mathbb{R}$ and projection $\boldsymbol{\pi} = \mathbf{I}_V$. Newton's method optimizes $\ell$ via weight updates

$$\mathbf{w}^{t+1} = \mathbf{w}^t - \boldsymbol{\xi}^t \quad \text{with} \quad \boldsymbol{\xi}^t = \eta^t \cdot \mathbf{H}^{-1}(\mathbf{w}^t) \cdot \nabla \ell(\mathbf{w}^t),$$

where $H_{ij}(\mathbf{w}) = \frac{\partial^2 \ell}{\partial w_i \partial w_j}(\mathbf{w})$ is the Hessian and $\eta^t > 0$.

**Lemma 5.** *If $\ell$ is strictly convex then Newton's method is safe, i.e. $\langle \mathbf{H}^{-1} \nabla \ell, \nabla \ell \rangle \geq 0$ for all $\mathbf{w}$.*

*Proof.* Factorize the Hessian at $\mathbf{w}^t$ as $\mathbf{H} = \mathbf{PDP}^\intercal$. If $\ell$ is strictly convex then $\mathbf{D} > 0$ and so

$$\langle \mathbf{H}^{-1} \nabla \ell, \nabla \ell \rangle = \langle \mathbf{D}^{-1} \mathbf{P}^\intercal \nabla \ell, \mathbf{P}^\intercal \nabla \ell \rangle \geq 0$$

as required. □

Two features are noteworthy: (i) the transform $\boldsymbol{\eta} = \mathbf{P}^\intercal \boldsymbol{\xi}$ *diagonalizes* the second-order Taylor expansion of $\ell$,

$$\text{compare } \ell(\mathbf{w} + \boldsymbol{\xi}) = \ell(\mathbf{w}) + \boldsymbol{\xi}^\intercal \cdot \nabla \ell + \frac{1}{2}\boldsymbol{\xi}^\intercal \mathbf{H} \boldsymbol{\xi}$$

$$\text{with } \ell(\mathbf{w} + \mathbf{P}\boldsymbol{\eta}) = \ell(\mathbf{w}) + \boldsymbol{\eta}^\intercal (\mathbf{P}^\intercal \nabla \ell) + \frac{1}{2}\boldsymbol{\eta}^\intercal \mathbf{D} \boldsymbol{\eta},$$

and (ii) the proof hinges on the *positivity* of $\mathbf{D}$. Sections A2 and A3 extend the approach to show the natural gradient (Amari, 1998) and mirror descent (Raskutti & Mukherjee, 2015) are safe using the Legendre transform.

### 3.3. Strongly-typed games are safe

We apply the lessons from the warmups to define a factorization of nonlinear functions.

**Definition 4.** *The functions $\{\ell_n : V \to \mathbb{R}\}_{n=1}^N$ simultaneously factorize if there is a triple*

$$\Big(\{\mathbf{P}_l\}_{l=1}^L, \{f_l : \mathbb{R}^{p_l} \to \mathbb{R}\}_{l=1}^L, \{g_n : \mathbb{R}^L \to \mathbb{R}\}_{n=1}^N\Big),$$

*satisfying*

$$\ell_n(\mathbf{w}) = g_n\Big(f_1(\mathbf{P}_1^\intercal \mathbf{w}), \ldots, f_L(\mathbf{P}_L^\intercal \mathbf{w})\Big) \quad \text{for all } n$$

*where $\mathbf{P}_l$ are $(D \times p_l)$-matrices whose columns jointly form an orthogonal basis of $V$ and $f_l$ and $g_n$ are differentiable, and the $g_n$'s **co-vary monotonically**: $\frac{\partial g_m}{\partial f_l} \frac{\partial g_n}{\partial f_l} \geq 0$.*

The projections $\tau_l = \mathbf{P}_l \mathbf{P}_l^\intercal$ define a type structure on $V$. Intuitively, the outputs $z_l = f_l(\mathbf{P}_l^\intercal \mathbf{w})$ are latent factors computed from the inputs $\mathbf{w}$ such that each $z_l$ is *independent* of the others – independence is enforced by the projections $\tau_l$. Monotonic covariation of the functions $g_n$ with respect to the factors $z_l$ plays the same role as positivity in two-player games and Newton's method.

**Definition 5.** *Game $(\mathcal{T}_V, \{\ell_n\}_{n=1}^N)$ is **strongly-typed** if the loss functions admit a simultaneous factorization whose projections $\{\tau_l = \mathbf{P}_l \mathbf{P}_l^\intercal\}_{l=1}^L$ **commute** with $\{\boldsymbol{\pi}_n\}_{n=1}^N$.*

**Theorem 2.** *Strongly-typed games are safe.*

*Proof.* Commutativity implies there is a basis $\{\mathbf{e}_i\}_{i=1}^D$ for $V$ that simultaneously diagonalizes the projections $\{\boldsymbol{\pi}_n\}$ and $\{\tau_l\}$. Express elements of $V$ as $(v_1, \ldots, v_D)$ in the basis. Safety then reduces to showing

$$\langle \boldsymbol{\pi}_m(\nabla \ell_m), \nabla \ell_n \rangle = \sum_{\{i : \boldsymbol{\pi}_m(\mathbf{e}_i) \neq 0\}} \frac{\partial g_m}{\partial v_i} \cdot \frac{\partial g_n}{\partial v_i} \geq 0.$$

Observe that $\frac{\partial f_k}{\partial v_i} \frac{\partial f_l}{\partial v_i} = 0$ if $k \neq l$ since $f_k$ and $f_l$ are functions of orthogonal coordinates. It follows that

$$\frac{\partial g_m}{\partial v_i} \cdot \frac{\partial g_n}{\partial v_i} = \left(\sum_{k=1}^L \frac{\partial g_m}{\partial f_k} \frac{\partial f_k}{\partial v_i}\right) \cdot \left(\sum_{l=1}^L \frac{\partial g_n}{\partial f_l} \frac{\partial f_l}{\partial v_i}\right)$$

$$= \sum_{l=1}^L \frac{\partial g_m}{\partial f_l} \frac{\partial g_n}{\partial f_l} \cdot \left(\frac{\partial f_l}{\partial v_i}\right)^2 \geq 0$$

since the $g_n$'s co-vary monotonically. □

Strong-typing is a sufficient but not necessary condition for safety. More general definitions can be proposed according to taste. Definition 5 is easy to check, covers the basic examples below, and incorporates the concrete intuition developed in the warmups.

### 3.4. Comparison with potential games

The proof of theorem 1 suggests that safe games are related to potential games (Monderer & Shapley, 1996). In our notation, a block game is a weighted potential game if there exists a potential function $\Phi$ and scalar weights $\alpha_n > 0$ satisfying

$$\ell_n(\mathbf{w}) - \ell_n(\mathbf{w} + \boldsymbol{\pi}_n \mathbf{v}) = \alpha_n \cdot \Big(\Phi(\mathbf{w}) - \Phi(\mathbf{w} + \boldsymbol{\pi}_n \mathbf{v})\Big)$$

for all $\mathbf{w}, \mathbf{v} \in V$ and $n \in [N]$.

We provide two counter-examples to show that strongly-typed games are distinct from potential games.

*Example* 5 (a strongly-typed game that is not a potential game). Let $\ell_1(\mathbf{x}, \mathbf{y}) = x_1 y_1 + 2 x_2 y_2$ and $\ell_2(\mathbf{x}, \mathbf{y}) = 3 x_1 y_1 + 4 x_2 y_2$. The block game with projections onto $\mathbf{x}$ and $\mathbf{y}$ is strongly-typed but is not a weighted potential game.

*Example* 6 (a potential game that is not safe). Let $\ell_1(x, y) = xy$ and $\ell_2(x, y) = xy - 9x$, with projections onto $x$ and $y$. The game is a potential game but is not safe because

$$\langle \boldsymbol{\pi}_1(\nabla \ell_1), \nabla \ell_2 \rangle = \langle (y, 0), (y - 9, x) \rangle = y^2 - 9y$$

can be negative.

## 4. Quadratic Games

Given a collection of $(D \times D)$-matrices $\{A^{(n)}\}_{n=1}^N$ and $D$-vectors $\{\mathbf{b}^{(n)}\}$ the corresponding **quadratic game** has loss functions

$$\ell_n(\mathbf{w}) = \frac{1}{2} \mathbf{w}^\intercal \mathbf{A}^{(n)} \mathbf{w} + \mathbf{w}^\intercal \mathbf{b}^{(n)}.$$

We assume the matrices $\mathbf{A}^{(n)}$ are symmetric without loss of generality.

### 4.1. Open quadratic games

In an **open quadratic game**, each player updates the entire joint action.

**Corollary 1.** *An open quadratic game is safe if there is an orthogonal $(D \times D)$-matrix $\mathbf{P}$, diagonal matrices $\mathbf{D}^{(n)}$ such that $\mathbf{D}^{(m)} \mathbf{D}^{(n)} \geq 0$, and $D$-vector $\mathbf{b}$ such that*

$$\mathbf{A}^{(n)} = \mathbf{P} \mathbf{D}^{(n)} \mathbf{P}^\intercal \quad \text{and} \quad \mathbf{b}^{(n)} = \mathbf{A}^{(n)} \mathbf{b}.$$

We derive corollaries 1 and 2 from theorem 2. Alternate, direct proofs are provided in appendix A4.

*Proof.* Let $f_i(x) = x(\frac{x}{2} - b_i)$ and $g_n(\mathbf{z}) = \sum_{i=1}^D d_i^{(n)} \cdot z_i$. Then

$$\ell_n(\mathbf{w}) = g_n\Big(f_1(\mathbf{p}_1^\intercal \mathbf{w}), \ldots, f_D(\mathbf{p}_D^\intercal \mathbf{w})\Big),$$

where $\mathbf{p}_i$ are the columns of $\mathbf{P}$, is strongly-typed. □

The Hessian of $\ell_n$ is $\mathbf{H}_{\ell_n} = \mathbf{A}^{(n)}$. The conditions of corollary 1 can be reformulated as (i) the Hessians of the losses commute $\mathbf{H}_{\ell_m}\mathbf{H}_{\ell_n} = \mathbf{H}_{\ell_n}\mathbf{H}_{\ell_m}$ for all $m$ and $n$, and (ii) the Newton steps for the losses coincide (when the Hessians are nonsingular):

$$\underbrace{\mathbf{H}_{\ell_n}^{-1}(\nabla \ell_n)}_{\text{Newton step}} = (\mathbf{A}^{(n)})^{-1}\mathbf{A}^{(n)}(\mathbf{w}-\mathbf{b}) = \mathbf{w}-\mathbf{b}.$$

**Example: Disentangling latent factors.** An important problem in machine learning is disentangling latent factors (Bengio, 2013). Basic methods for tackling the problem include PCA, canonical correlation analysis (CCA) and independent component analysis (ICA). We show how the factorization in corollary 1 can arise "in nature" as a variant of blind source separation.

Suppose a signal on $D$ channels is recorded for $T$ timepoints giving $(D \times T)$-matrix $\mathbf{X}$. Assume the observations combine $L$ *independent* latent signals: $\mathbf{X} = \mathbf{MS}$ where $\mathbf{S}$ is an $(L \times T)$-matrix representing the latent signal and $\mathbf{M}$ is a mixing matrix.

Blind source separation is concerned with recovering the latent signals. The covariance of the signal is $\mathbf{A} = \mathbf{XX}^\intercal$. Factorize $\mathbf{A} = \mathbf{PDP}^\intercal$ and let $\tilde{\mathbf{S}} = \mathbf{P}^\intercal \mathbf{X}$. Although this may not recover the original signal, i.e. $\tilde{\mathbf{S}} \neq \mathbf{S}$ in general, it does disentangle $\mathbf{X}$ into uncorrelated factors:

$$\tilde{\mathbf{S}}\tilde{\mathbf{S}}^\intercal = \mathbf{P}^\intercal \mathbf{XX}^\intercal \mathbf{P} = \mathbf{D}.$$

Finally, recall that finding the first principal component can be formulated as the constrained maximization problem:

$$\underset{\{\mathbf{w}:\|\mathbf{w}\|_2=1\}}{\operatorname{argmax}} \mathbf{w}^\intercal \mathbf{A}\mathbf{w}.$$

Now suppose there are $N$ sets of observations $\mathbf{X}^{(1)}, \ldots, \mathbf{X}^{(N)}$ generated by a single orthogonal mixing matrix acting on different sets of (potentially rescaled) latent signals: $\mathbf{X}^{(n)} = \mathbf{PS}^{(n)}$. Finding the first principle components of the signals reduces to solving the constrained optimization problems

$$\left\{\underset{\{\mathbf{w}:\|\mathbf{w}\|_2=1\}}{\operatorname{argmax}} \mathbf{w}^\intercal \mathbf{X}^{(n)}(\mathbf{X}^{(n)})^\intercal \mathbf{w}\right\}_{n=1}^N \quad (3)$$

Corollary 1 implies that (3) is a safe. Note the corollary implies the optimization problems have compatible gradients, not that they share a common solution. In general there are many Nash equilibria, analogous to example 3.

**Quadratic games and linear regression.** The blind source separation example assumes that the linear terms $\mathbf{b}^{(n)}$ in the loss are zero. If the linear term is nonzero then linear regression is a special case of minimizing the quadratic loss. Safety then relates to searching for weights that simultaneously solve linear regression problems on multiple datasets.

### 4.2. Block Quadratic Games

The **block quadratic game** has losses as above; however the action space decomposes into $(\mathbf{w}_1, \ldots, \mathbf{w}_N)$ with corresponding projections. Block decompose the components of the loss as

$$\mathbf{A}^{(n)} = \begin{pmatrix} \mathbf{A}_{11}^{(n)} & \cdots & \mathbf{A}_{1N}^{(n)} \\ \vdots & & \vdots \\ \mathbf{A}_{N1}^{(n)} & \cdots & \mathbf{A}_{NN}^{(n)} \end{pmatrix} \text{ and } \mathbf{b}^{(n)} = \begin{pmatrix} \mathbf{b}_1^{(n)} \\ \vdots \\ \mathbf{b}_N^{(n)} \end{pmatrix}.$$

**Corollary 2.** *A block quadratic game is safe if there are:*

*(i) $(D \times D)$-orthogonal $\mathbf{P}$ with $\mathbf{P}_{mn} = 0$ for $m \neq n$;*

*(ii) $(D \times L)$ matrix $\mathbf{R}$ with $\mathbf{R}_{n\bullet}$ diagonal for all $n$;*

*(iii) diagonal $(L \times L)$-matrices $\mathbf{D}^{(n)}$ with $\mathbf{D}^{(m)}\mathbf{D}^{(n)} \geq 0$;*

*(iv) and a $D$-vector $\mathbf{b}$*

*such that* $\mathbf{A}^{(n)} = \mathbf{PRD}^{(n)}\mathbf{R}^\intercal \mathbf{P}^\intercal$ *and*
$$\mathbf{b}^{(n)} = \mathbf{A}^{(n)}\mathbf{b} \quad \text{for all } n.$$

The notation $\mathbf{P}_{mn}$ and $\mathbf{R}_{\bullet n}$ refers to blocks in the rows and columns of $\mathbf{P}$ and columns of $\mathbf{R}$.

*Proof.* Let $\mathbf{p}_i$ denote the columns of $\mathbf{P}$ and $g_n(\mathbf{z}) = \sum_{l=1}^L d_l^{(n)} \cdot z_l$. Given $l$, construct $\mathbf{P}_l$ by concatenating the columns $\mathbf{p}_i$ of $\mathbf{P}$ for which the corresponding entries of $\mathbf{R}_{il}$ are nonzero and let $\mathbf{r}_l$ be the vector containing the nonzero entries of $R_{\bullet l}$. Define $f_l(\mathbf{x}_l) = \mathbf{r}_l^\intercal(\frac{\mathbf{x}_l}{2} - \mathbf{b}_l) \cdot (\mathbf{r}_l^\intercal \mathbf{x}_l)$. Then

$$\ell_n(\mathbf{w}) = g_n\Big(f_1(\mathbf{P}_1^\intercal \mathbf{w}), \ldots, f_L(\mathbf{P}_L^\intercal \mathbf{w})\Big).$$

It is an exercise to check the game is strongly-typed. □

**Example: Disentangling latent factors.** We continue the discussion of blind source separation and safety. Suppose that the mixing matrix decomposes into blocks

$$\mathbf{M} = \begin{pmatrix} \mathbf{M}_{1\bullet} \\ \vdots \\ \mathbf{M}_{N\bullet} \end{pmatrix}$$

The blocks generate *multiple views* on a single latent signal, (Kakade & Foster, 2007; McWilliams et al., 2013; Benton et al., 2017). The $n^{\text{th}}$ view is $\mathbf{M}_{n\bullet}\mathbf{S}$.

As in the example in section 4.1, now suppose there are $N$ sets of observed signals generated from $N$ sets of latent signals. Each agent attempts to find the principal component specific to its view on its set of observations. Corollary 2 implies that the problems

$$\left\{\underset{\{\mathbf{w}_n:\|\mathbf{w}_n\|_2=1\}}{\operatorname{argmax}} \mathbf{w}^\intercal \mathbf{X}^{(n)}(\mathbf{X}^{(n)})^\intercal \mathbf{w}\right\}_{n=1}^N$$

can be safely optimized using gradient descent if the mixing matrix has the block form

$$\mathbf{M}_{n\bullet} = \mathbf{P}_{nn} \cdot \mathbf{R}_{n\bullet}$$

where $\mathbf{P}_{nn}$ is orthogonal and $\mathbf{R}_{n\bullet}$ is diagonal. In other words, if the views are generated by rescaling and changing-the-basis of the latent signals.

The open and block settings share a common theme: Safe disentangling requires observed signals that are generated by a single (structured) mixing process applied to (arbitrary) sets of independent latent signals. The same phenomenon arises in multi-player games, resulting in tensor decompositions that generalize ICA.

## 5. Multi-Player Games and Tensor-SVD

A classic $N$-player strategic game consists in finite action-sets $A_n$ and losses $\ell_n : A = \prod_{n=1}^N A_n \to \mathbb{R}$. Enumerate the elements of each set as $A_n = [D_n]$, and encode the losses as $(D_1, \ldots, D_N)$-tensors

$$\mathcal{A}_n[\alpha_1, \ldots, \alpha_N] := \ell_n(\alpha_1, \ldots, \alpha_N) \text{ where } \alpha_n \in [D_n].$$

Given a collection of $N$ such tensors, define the corresponding **multilinear game**[1] as

$$\ell_n(\mathbf{w}_1, \ldots, \mathbf{w}_N) = \mathcal{A}_n \times_1 \mathbf{w}_1 \times \cdots \times_N \mathbf{w}_N$$
$$:= \sum_{\alpha_1, \ldots, \alpha_N = 1}^{D_1, \ldots, D_N} \mathcal{A}[\alpha_1, \ldots, \alpha_N] \cdot \mathbf{w}_1[\alpha_1] \cdots \mathbf{w}_N[\alpha_N].$$

The classic example is when actions are drawn from the $D_n$-simplex $\triangle^{D_n} = \{\mathbf{w}_n \in \mathbb{R}^{D_n} : \sum_{\alpha=1}^{D_n} \mathbf{w}_n[\alpha] = 1 \text{ and } \mathbf{w}_n[\alpha] \geq 0 \text{ for all } \alpha\}$.

We now recall the orthogonal tensor decomposition or tensor SVD (Zhang & Golub, 2001; Chen & Saad, 2009). A tensor admits a tensor-SVD if it can be written in the form

$$\mathcal{A} = \sum_{l=1}^L d_l \cdot \mathbf{u}_l^1 \otimes \cdots \otimes \mathbf{u}_l^N = \mathcal{D} \times_1 \mathbf{U}^1 \times \cdots \times_N \mathbf{U}^N$$

where $\mathbf{U}^n$ is a $(D_n \times L)$-matrix with orthogonal columns and $\mathcal{D}$ is a diagonal $(L, \ldots, L)$-tensor.

**Corollary 3.** *A multilinear game is safe if it admits a simultaneous tensor-SVD*

$$\mathcal{A}^{(n)} = \mathcal{D}^{(n)} \times_1 \mathbf{U}_1 \times \cdots \times_N \mathbf{U}^N$$

*where the diagonals have the same sign coordinatewise.*

---
[1] We use the $n$-mode product notation $\times_n$, see de Lathauwer et al. (2000).

*Proof.* Let $g_n(\mathbf{z}) = \sum_{l=1}^L d_l^{(n)} z_l$ and $f_l(\mathbf{x}) = \prod_n x_n$. Define $\mathbf{P}_l$ as the $(D \times N)$-matrix whose $n^{\text{th}}$ column is $\mathbf{u}_l^n$ in the block of rows corresponding to $\mathbf{w}_n$ and zero elsewhere. Then

$$\ell_n(\mathbf{w}) = g_n\big(f_1(\mathbf{P}_1^\intercal \mathbf{w}), \ldots, f_L(\mathbf{P}_L^\intercal \mathbf{w})\big)$$

and the game is strongly-typed. □

Not all tensors admit a tensor-SVD. However, all tensors do admit a higher-order SVD (de Lathauwer et al., 2000). Section A5 explains why simultaneous HOSVD does not guarantee safety and the stronger tensor-SVD is required.

**Example: Disentangling latent factors** Suppose $\mathbf{S}$ is a latent signal with independent non-Gaussian coordinates. We observe $\mathbf{X} = \mathbf{PS} + \boldsymbol{\epsilon}$ where $\mathbf{P}$ is a $(D \times L)$ mixing matrix and $\boldsymbol{\epsilon}$ is Gaussian noise. By whitening the signal as a preprocessing step, one can ensure the columns of $\mathbf{P}$ are orthogonal. ICA recovers $\mathbf{S}$ from the cumulants of $\mathbf{X}$, see Hyvärinen et al. (2001). The main insight is that the $4^{\text{th}}$-order cumulant tensor admits a tensor-SVD:

$$\mathcal{A}[i, j, k, l] = \text{cum}(x_i, x_j, x_k, x_l)$$
$$= \sum_{o,p,q,r} P_{io} P_{jp} P_{kq} P_{lr} \cdot \text{cum}(s_o, s_p, s_q, s_r)$$
$$= \sum_r P_{ir} P_{jr} P_{kr} P_{lr} \cdot \text{kurt}(s_r)$$

since $\text{cum}(s_o, s_p, s_q, s_r) = 0$ unless $o = p = s = r$ because the signals are independent. The expression can be written $\mathcal{A} = \mathcal{K} \times_1 \mathbf{P} \times_2 \mathbf{P} \times_3 \mathbf{P} \times_4 \mathbf{P}$ where diagonal tensor $\mathcal{K}$ specifies the kurtosis of the latent signal. In other words, computing the tensor-SVD recovers the mixing matrix and allows to recover the latent signal up to basic symmetries.

Following the same prescription as the examples above, if there are $N$ sets of observations generated from $N$ latent signals by the same mixing matrix, then the resulting cumulant tensors satisfy corollary 3.

## 6. Biologically Plausible Backpropagation

Our ultimate goal is to apply strong-typing to safely optimize neural nets with multiple loss functions (Marblestone et al., 2016). Doing so requires constructing variants of backprop that allow the propagation of multiple error signals. First steps in this direction have been taken with biologically plausible models of backprop that introduce additional degrees of freedom into the algorithm.

**Feedback alignment** is a recent algorithm with comparable empirical performance to backprop. It is also more biologically plausible since it loosens backprop's requirement that forward- and back- propagating weights are sym-metric (Lillicrap et al., 2016). The main theoretical result of the paper, see their supplementary information, is

**Theorem.** *Let $\boldsymbol{\delta}_{BP} = \mathbf{W}^\intercal \mathbf{e}$ denote the error backpropagated one layer of the neural network. Under certain conditions, the error signal computed by feedback alignment, $\boldsymbol{\delta}_{FA} = B\mathbf{e}$, satisfies*

$$\boldsymbol{\delta}_{FA} = \alpha \cdot \mathbf{W}^\dagger \mathbf{e} \text{ where } \alpha > 0$$

*and $\mathbf{W}^\dagger$ is the pseudoinverse of $\mathbf{W}$.*

*Proof.* See theorem 2 of Lillicrap et al. (2016). □

**Corollary 4.** *Under the same conditions, feedback alignment is safe.*

*Proof.* We require to check $\langle \boldsymbol{\delta}_{FA}, \boldsymbol{\delta}_{BP} \rangle \geq 0$. Applying the theorem obtains

$$\langle \boldsymbol{\delta}_{FA}, \boldsymbol{\delta}_{BP} \rangle = \alpha \cdot \langle \mathbf{W}^\dagger \mathbf{e}, \mathbf{W}^\intercal \mathbf{e} \rangle = \alpha \cdot \langle \mathbf{W} \mathbf{W}^\dagger \mathbf{e}, \mathbf{e} \rangle.$$

Observe that $\mathbf{W}\mathbf{W}^\dagger$ is an orthogonal projection by standard properties of the pseudoinverse so

$$\langle \boldsymbol{\delta}_{FA}, \boldsymbol{\delta}_{BP} \rangle = \alpha \cdot \langle \mathbf{W}\mathbf{W}^\dagger \mathbf{e}, \mathbf{W}\mathbf{W}^\dagger \mathbf{e} \rangle \geq 0$$

as required. □

In fact, Lillicrap et al. (2016) provide experimental and theoretical evidence that feedback alignment learns to align the feedforward weights with the pseudoinverse of the backconnections. In other words, they argue that feedback alignment *learns safe gradients*.

Another variant of backprop is **kickback**, which loosens backprop's requirement that there are distinct forward- and backward signals (Balduzzi et al., 2015). Kickback truncates backprop's error signals so that the network learns from just the feedforward sweep together with scalar error signals. One of the main results of Balduzzi et al. (2015) is that kickback is safe, see section A6.

## 7. Conclusion

Backprop provides a general-purpose tool to train configurations of differentiable modules that **share a single objective**. However, effectively training populations of neural networks on potentially conflicting tasks, such that they automatically exploit synergies and avoid damaging incompatibilities (such as unlearning old features that are not useful on a new task) requires fundamentally new ideas.

A key piece of the puzzle is to develop *type systems* that can be used to (i) guarantee when certain optimizations can be safely performed jointly and (ii) flag potential conflicts so that the incompatible optimization problems can be separated. The paper provides a first step in this direction.

From a different perspective, convex methods have played an enormous role in optimization yet their relevance to deep learning is limited. The approach to strong-typing developed here is inspired by and extends certain features of convexity. One of the goals of this paper is to carve out some of the key concepts underlying convex geometry and reassemble them into a more flexible, but still powerful framework. The proposed definition of strong-typing should be considered a first and far from final attempt.

A large class of natural examples is generated by imposing strong-typing on simple quadratic and multilinear games. It turns out that, in these settings, strong-typing yields the same matrix and tensor decompositions that arise in blind source separation and independent component analysis, where multiple latent signals are mixed by the same structured process. An important future direction is to disentangle nonlinear latent factors.

**Strong-typing and safety in neural nets.** We conclude by discussing the relevance of the framework to neural networks. Firstly, neural nets and strong-typing have many of the same ingredients: neural nets combine linear algebra (matrix multiplications and convolutions) with monotonic functions (sigmoids, tanhs, rectifiers, and max-pooling amongst others). Rectifiers and sigmoids have the additional feature that their outputs are always positive.

Secondly, there is a deeper connection between rectifiers and strong-typing. **Rectifiers are orthogonal projections on weights**: $\rho(\mathbf{W}^\intercal \mathbf{x})$ zeroes out the columns $\mathbf{w}_l$ of $\mathbf{W}$ for which $\mathbf{w}_l^\intercal \mathbf{x} \leq 0$. Rectifiers are more sophisticated projections than we have previously considered because they are context-dependent. The columns that are zeroed out depend on $\mathbf{W}$ and $\mathbf{x}$: the rectifier-projection takes $\mathbf{W}$ and $\mathbf{x}$ as parameters, compare remarks 1 and 2 in the appendix. Representation learning in rectifier networks can thus be recast as learning parameterized type structures. An interesting future direction is to consider tensor-switching networks (Tsai et al., 2016), which decouple a neuron's decision to activate from the information it passes along (for a rectifier, both depend on $\mathbf{W}^\intercal \mathbf{x}$).

Finally, it has long been known that the brain does not use backprop (Crick, 1989). One possibility is that backprop is the optimal deep learning algorithm which, unfortunately, evolution failed to stumble upon. Another is that there are *evolutionary advantages to not using backpropagation*. For example, it has been argued that the brain optimizes multiple loss functions (Marblestone et al., 2016). Does jointly optimizing or satisficing multiple objectives require learning mechanisms with more degrees of freedom than backprop (Balduzzi et al., 2015; Lillicrap et al., 2016)? Safety and strong-typing provide the tools needed to frame and investigate the question.

## Acknowledgements

I am grateful to Stephen Marsland and James Benn for useful discussions.

# APPENDIX

## A1. Proof of Theorem 1

For completeness we recall the notion of Nash equilibrium, generalized to our setup:

**Definition A1.** *The point $\mathbf{w}^* \in \mathcal{H}$ is a **Nash equilibrium** if for all players $n$ and for all $\mathbf{w}$ satisfying $\mathbf{w}^* + \boldsymbol{\pi}_n(\mathbf{w}) \in \mathcal{H}$ it holds that*

$$\ell_n(\mathbf{w}^*) \leq \ell_n(\mathbf{w}^* + \boldsymbol{\pi}_n(\mathbf{w})).$$

This section completes theorem 1 when gradient descent converges to the boundary of the feasible set. To ensure actions stay in the feasible set we use gradient updates of the form

$$\mathbf{w}^{t+1} = \texttt{Proj}_{\mathcal{H}}(\mathbf{w}^t - \boldsymbol{\pi}_{\rho(n)}(\boldsymbol{\xi}_n^t)) \quad \text{where}$$
$$\texttt{Proj}_{\mathcal{H}}(\tilde{\mathbf{x}}) = \underset{\mathbf{x} \in \mathcal{H}}{\operatorname{argmin}} \|\mathbf{x} - \tilde{\mathbf{x}}\|_2.$$

*Proof.* A point on the boundary of $\mathcal{H}$ is a critical point of $\Phi$ if the gradient is zero or points perpendicularly out the constraint set. Since $\langle \boldsymbol{\pi}_m(\nabla \Phi), \nabla \ell_m \rangle \geq \alpha_m \|\boldsymbol{\pi}_m(\nabla \ell_m)\|_2^2 \geq 0$ for all players $m$, it follows that gradient descent by the players, with step sizes decaying as $\frac{1}{\sqrt{T}}$, will converge to a critical point of $\Phi$. Critical points of $\Phi$ are Nash equilibria of the game by convexity of the losses as a function of their players' actions. $\square$

## A2. The natural gradient is safe

The natural gradient was introduced in Amari (1998) and is widely used in machine learning.

**Theorem.** *The direction of steepest descent at $\ell(\mathbf{w}^t)$ on Riemannian manifold $(\mathbb{R}^n, \mathbf{G})$ is*

$$\boldsymbol{\xi}^t \propto \mathbf{G}^{-1}(\mathbf{w}^t) \cdot \nabla \ell(\mathbf{w}^t).$$

*The **natural gradient** is the direction $\mathbf{G}^{-1}(\mathbf{w}^t) \cdot \nabla \ell(\mathbf{w}^t)$.*

*Proof.* We follow Amari (1998). The problem reduces to the constrained minimization

$$\underset{\{\boldsymbol{\xi}: \sum G_{ij}(\mathbf{w}^t)\xi_i\xi_j = 1\}}{\operatorname{argmin}} \left[ \ell(\mathbf{w}^t) + \epsilon \nabla \ell(\mathbf{w}^t)^\intercal \boldsymbol{\xi} \right]$$

which can be rewritten as

$$\underset{\boldsymbol{\xi}}{\operatorname{argmin}} \left[ \nabla \ell(\mathbf{w}^t)^\intercal \boldsymbol{\xi} - \lambda \boldsymbol{\xi}^\intercal \mathbf{G}(\mathbf{w}^t) \boldsymbol{\xi} \right]$$

with solution

$$\boldsymbol{\xi}^t \propto \mathbf{G}^{-1}(\mathbf{w}^t) \nabla \ell(\mathbf{w}^t)$$

as required. $\square$

**Corollary A1.** *The natural gradient is safe.*

The proof is obvious. We include it to highlight the role of latent types, diagonalization and positivity.

*Proof.* We are required to show that

$$\langle \mathbf{G}^{-1}(\mathbf{w}) \nabla \ell(\mathbf{w}), \nabla \ell(\mathbf{w}) \rangle \geq 0 \text{ for all } \mathbf{w}. \quad \text{(A1)}$$

The metric is symmetric positive definite, and so admits factorization

$$\mathbf{G}(\mathbf{w}^t) = \mathbf{P}(\mathbf{w}) \cdot \mathbf{D}(\mathbf{w}) \cdot \mathbf{P}^\intercal(\mathbf{w})$$

where $\mathbf{P}(\mathbf{w})$ is an orthogonal matrix and $\mathbf{D}(\mathbf{w})$ strictly positive diagonal for all $\mathbf{w}$. After setting $\tilde{\nabla}\ell(\mathbf{w}) := \mathbf{P}^\intercal(\mathbf{w}) \nabla \ell(\mathbf{w})$, the condition in (A1) can be rewritten as

$$\langle \mathbf{D}^{-1}(\mathbf{w})\tilde{\nabla}\ell(\mathbf{w}), \tilde{\nabla}\ell(\mathbf{w}) \rangle \geq 0$$

which clearly holds. $\square$

*Remark* 1 (parametric types). The columns $\mathbf{p}_i(\mathbf{w})$ of $\mathbf{P}(\mathbf{w})$ define a *parametric family* of latent type systems $\boldsymbol{\tau}_i(\mathbf{w}) = \mathbf{p}_i(\mathbf{w}) \cdot \mathbf{p}_i^\intercal(\mathbf{w})$ that is parametrized by $\mathbf{w}$.

## A3. Mirror descent is safe

We show that mirror descent is safe. The fact that mirror descent is well-behaved is far from new; there is an extensive literature analyzing its convergence properties (Bubeck, 2015). Our purpose is to highlight the type structures underlying convex duality and mirror descent.

The approach generalizes the analysis of Newton's method. The role played by transform that diagonalizes the Hessian in section 3.2 is taken over by the Legendre transform here.

**Convex duality.** The exposition closely follows Amari (2009). Consider the manifold $M = \mathbb{R}^D$ with coordinate system $\mathbf{w} = (w_1, \ldots, w_D)$. Let $\psi : M \to \mathbb{R}$ be a twice-differentiable strictly convex function. It follows that the Hessian of $\psi$,

$$g_{ij}(\mathbf{w}) = \partial_i \partial_j \psi(\mathbf{w})$$

is a positive definite matrix for all $\mathbf{w}$ which defines a *Riemannian metric* on the manifold $M$. Concretely, the distance between infinitesimally close points $\mathbf{w}$ and $\mathbf{w} + d\mathbf{w}$ is

$$ds^2 = \sum_{ij} g_{ij}(\mathbf{w}) dw^i dw^j.$$

Define the *dual* coordinate system

$$\boldsymbol{\theta} = \nabla \psi(\mathbf{w}), \quad \text{i.e. } \theta_i = \frac{\partial \psi}{\partial w_i}(\mathbf{w}).$$

Recall that the **Legendre transform** of $\psi$,

$$\psi^*(\boldsymbol{\theta}) = \max_{\mathbf{w}} \{\mathbf{w}^\intercal \boldsymbol{\theta} - \psi(\mathbf{w})\},$$

satisfies the relation

$$\psi(\mathbf{w}) + \psi^*(\boldsymbol{\theta}) - \mathbf{w}^\intercal \boldsymbol{\theta} = 0,$$

which allows to recover the original coordinates from the dual system:

$$\mathbf{w} = \nabla \psi^*(\boldsymbol{\theta}) \quad \text{i.e. } w_i = \frac{\partial \psi^*}{\partial \theta_i}(\boldsymbol{\theta}).$$

Define the *dual metric*

$$g^{ij}(\boldsymbol{\theta}) = \frac{\partial^2}{\partial \theta_i \partial \theta_j} \psi^*(\boldsymbol{\theta}).$$

**Theorem.** *The metrics $g_{ij}(\mathbf{w})$ and $\mathbf{g}^{ij}(\boldsymbol{\theta})$ are inverse. That is*

$$d\boldsymbol{\theta} = \mathbf{G}(\mathbf{w})d\mathbf{w} \quad \text{and} \quad d\mathbf{w} = \mathbf{G}^{-1}(\boldsymbol{\theta})d\boldsymbol{\theta}.$$

*Proof.* See Amari (2009). □

*Remark* 2 (parametric types). Convex duality can thus be rephrased as a relationship between two parametric families of types on the tangent spaces to the manifold that is encoded in the Riemannian metric and its inverse.

**Mirror descent is safe.** We show that mirror descent is safe by applying convex duality following Raskutti & Mukherjee (2015). Let $\psi$ be a strictly convex twice differentiable function. The Bregman divergence of $\psi$ is $D_\psi(\mathbf{v}, \mathbf{w}) = \psi(\mathbf{v}) - \psi(\mathbf{w}) - \langle \nabla \psi(\mathbf{w}), \mathbf{v} - \mathbf{w} \rangle$.

**Theorem.** *Given loss function $\ell$ and strictly convex twice differentiable $\psi$, both defined on $M = \mathbb{R}^D$, the mirror descent step*

$$\mathbf{w}^{t+1} = \operatorname*{argmin}_{\mathbf{w}} \left\{ \mathbf{w}^\intercal \nabla \ell(\mathbf{w}^t) + \frac{1}{\eta^t} D_\psi(\mathbf{w}, \mathbf{w}^t) \right\},$$

*is equivalent to the natural gradient descent step*

$$\boldsymbol{\theta}^{t+1} = \boldsymbol{\theta}^t - \eta^t \cdot \left[ \nabla^2 \psi^*(\boldsymbol{\theta}^t) \right]^{-1} \cdot \nabla \ell(\boldsymbol{\theta}^t).$$

*in the dual coordinate system.*

*Proof.* We sketch the proof in Raskutti & Mukherjee (2015), which should be consulted for details. Computing the minimizer in mirror descent by differentiating shows that the mirror descent update is equivalent, in dual coordinates, to

$$\boldsymbol{\theta}^{t+1} = \boldsymbol{\theta}^t - \eta^t \cdot \nabla_{\mathbf{w}} \ell(\nabla \psi^*(\boldsymbol{\theta}^t))$$

Applying the chain rule obtains

$$\nabla_{\mathbf{w}} \ell(\nabla \psi^*(\boldsymbol{\theta}^t)) = \left[ \nabla^2 \psi^*(\boldsymbol{\theta}^t) \right]^{-1} \cdot \nabla_{\boldsymbol{\theta}} \ell(\nabla \psi^*(\boldsymbol{\theta}^t))$$

so that mirror descent can be rewritten as

$$\boldsymbol{\theta}^{t+1} = \boldsymbol{\theta}^t - \eta^t \cdot \left[ \nabla^2 \psi^*(\boldsymbol{\theta}^t) \right]^{-1} \cdot \nabla_{\boldsymbol{\theta}} \ell(\nabla \psi^*(\boldsymbol{\theta}^t))$$

as required. □

**Corollary A2.** *If $\psi$ is twice-differentiable and strictly convex then mirror descent is safe.*

*Proof.* Combine corollary A1 with the equivalence of mirror descent and the natural gradient. □

### A4. Direct proofs of corollaries 1 and 2

**Direct proof of corollary 1.**

*Proof.* Safety requires

$$\langle \nabla \ell_n, \nabla \ell_m \rangle = (\mathbf{A}^{(n)} \mathbf{w} - \mathbf{b}^{(n)})^\intercal (\mathbf{A}^{(m)} \mathbf{w} - \mathbf{b}^{(m)}) \geq 0.$$

Plugging in the assumptions yields

$$\langle \nabla \ell_n, \nabla \ell_m \rangle = (\mathbf{w} - \mathbf{b})^\intercal \mathbf{P} \mathbf{D}^{(n)} \mathbf{D}^{(m)} \mathbf{P}^\intercal (\mathbf{w} - \mathbf{b}) \geq 0$$

which is positive by inspection. □

**Direct proof of corollary 2.**

*Proof.* The gradient and projected gradient are

$$\nabla \ell_n = \mathbf{w}^\intercal \mathbf{A}^{(n)} + \mathbf{b}^{(n)} \text{ and}$$
$$\boldsymbol{\pi}_n(\nabla \ell_n) = \mathbf{w}^\intercal \mathbf{A}^{(n)}_{\bullet n} + \mathbf{b}^{(n)}_n$$

After imposing $\mathbf{b}^{(n)} = \mathbf{A}^{(n)}\mathbf{b}$, safety requires that

$$(\mathbf{w} - \mathbf{b})^\intercal \mathbf{A}^{(m)}_{\bullet m} \mathbf{A}^{(n)}_{m \bullet}(\mathbf{w} - \mathbf{b}) \geq 0 \text{ for all } \mathbf{w}, m \text{ and } n.$$

Applying the remaining conditions and setting $\mathbf{x} = \mathbf{R}^\intercal \mathbf{P}^\intercal (\mathbf{w} - \mathbf{b})$ obtains

$$\mathbf{x}^\intercal \mathbf{D}^{(m)} \mathbf{R}^\intercal_{\bullet m} \mathbf{R}_{m \bullet} \mathbf{D}^{(n)} \mathbf{x} \geq 0$$

Since $\mathbf{R}_{m\bullet}$ is diagonal, it follows that $\mathbf{D}^{(m)} \mathbf{R}^\intercal_{\bullet m} \mathbf{R}_{m\bullet} \mathbf{D}^{(n)}$ is a product of diagonal matrices. The expression is positive since $\mathbf{D}^{(m)} \mathbf{D}^{(n)} \geq 0$ by assumption. □

## A5. Higher-order SVD

Any $N$-tensor admits a higher-order SVD (HOSVD), whereas not every tensor admits a tensor-SVD. In this section we recall the concept of HOSVD and show why simultaneous HOSVD is not sufficient to guarantee safety. The section relies heavily on notation taken from de Lathauwer et al. (2000) which the reader is encouraged to consult for details and context.

**Definition A2** (matricization). *The **matricization** of tensor $\mathcal{A}$ over its $n^{th}$ mode, denoted $\mathbf{A}_{(n)}$ is an $D_n \times (D_{n+1} D_{n+2} \cdots D_N D_1 D_2 \cdots D_{n-1})$-matrix that contains the element $\mathcal{A}[\alpha_1, \ldots, \alpha_N]$ at the position with row number $\alpha_n$ and column number*

$$(\alpha_{n+1} - 1) D_{n+2} D_{n+3} \cdots D_N D_1 \cdots D_{n-1}$$
$$+ (\alpha_{n+2} - 1) D_{n+3} D_{n+4} \cdots D_N D_1 \cdots D_{n-1} + \cdots$$
$$+ (\alpha_N - 1) D_1 D_2 \cdots D_{n-1} + (\alpha_1 - 1) D_2 D_3 \cdots D_{n-1}$$
$$+ (\alpha_2 - 1) D_3 D_4 \cdots D_{n-1} + \cdots + \alpha_{n-1}$$

Let us recall the notion of higher-order SVD (HOSVD) from de Lathauwer et al. (2000).

**Theorem.** *Every $(I_1, \ldots, I_N)$-tensor $\mathcal{A}$ can be written as a product*

$$\mathcal{A} = \mathcal{S} \times_1 \mathbf{U}^1 \times_2 \mathbf{U}^2 \cdots \times_N \mathbf{U}^N$$

*in which*

1. $\mathbf{U}^i$ *is a unitary $(I_i \times I_i)$ matrix*

2. $\mathcal{S}$ *is a $(I_1 \times I_2 \cdots I_N)$-tensor of which the subtensors $\mathcal{S}_{i_n = \alpha}$, obtained by fixing the $n^{th}$ index to $\alpha$ have the properties of*

    (a) *all-orthogonality: two subtensors $\mathcal{S}_{i_n=\alpha}$ and $\mathcal{S}_{i_n=\beta}$ are orthogonal for all possible values of $n$, $\alpha$ and $\beta$ subject to $\alpha \neq \beta$:*

    $$\langle \mathcal{S}_{i_n=\alpha}, \mathcal{S}_{i_n=\beta} \rangle = 0 \quad \text{when} \quad \alpha \neq \beta$$

    (b) *ordering:*

    $$\|\mathcal{S}_{i_n=1}\| \geq \|\mathcal{S}_{i_n=2}\| \geq \cdots \geq \|\mathcal{S}_{i_n=I_n}\| \geq 0$$

    *for all possible values of $n$.*

    *The Frobenius-norms $\|\mathcal{S}_{i_n=i}\|$, symbolized by $\sigma_i^{(n)}$ are $n$-mode singular values of $\mathcal{A}$ and the vector $\mathbf{U}_i^{(n)}$ is an $i^{th}$ $n$-mode singular vector.*

*Proof.* See de Lathauwer et al. (2000). □

An important property of the HOSVD is as follows. Let $\mathcal{A} = \mathcal{S} \times_1 \mathbf{U}^1 \times \cdots \times_N \mathbf{U}^N$ be the HOSVD of $\mathcal{A}$. Then the matricization is

$$\mathbf{A}_{(n)} = \mathbf{U}^n \mathbf{\Sigma}^n (\mathbf{V}^n)^\intercal$$

is an SVD of $\mathbf{A}_{(n)}$, where the diagonal matrix $\mathbf{\Sigma}^n \in \mathbb{R}^{D_n \times D_n}$ and the columnwise orthonormal matrix $\mathbf{V}^n \in \mathbb{R}^{D_{n+1} D_{n+2} \cdots D_N D_1 D_2 \cdots D_{n-1} \times D_n}$ are defined as

$$\mathbf{\Sigma}^n := \text{diag}(\sigma_1^n, \ldots, \sigma_{D_n}^n),$$
$$(\mathbf{V}^n)^\intercal := \tilde{\mathbf{S}}_{(n)} \left( \mathbf{U}^{n+1} \otimes \cdots \mathbf{U}^N \otimes \mathbf{U}^1 \otimes \cdots \mathbf{U}^{n-1} \right)^\intercal$$

where $\tilde{\mathbf{S}}^n$ is a normalized version of the matricization $\mathbf{S}_{(n)}$ of $\mathcal{S}$, with rows rescaled to unit-length: $\mathbf{S}_{(n)} = \mathbf{\Sigma}^n \cdot \tilde{\mathbf{S}}^n$.

### A5.1. Discussion of safety and HOSVD

To somewhat reduce the proliferation of subscripts and superscripts, we work with two tensors denoted $\mathcal{A}$ and $\mathcal{B}$, which stand in for the tensors of two players $\mathcal{A}^{(m)}$ and $\mathcal{A}^{(n)}$ in an $N$-player game.

**Lemma A1.** *Let $\mathcal{A}[\mathbf{w}_{\hat{n}}]$ be the $D_n$-vector*

$$\mathcal{A}[\mathbf{w}_{\hat{n}}] := \mathcal{A} \times_1 \mathbf{w}_1 \times \cdots \times_{n-1} \mathbf{w}_{n-1} \times_{n+1} \mathbf{w}_{n+1} \times \cdots \times_N \mathbf{w}_N$$

*where note the term $\mathbf{w}_n$ is omitted from the expression, and similarly for $\mathcal{B}$. Then*

$$\langle \boldsymbol{\pi}_n \nabla \ell_{\mathcal{A}}, \nabla \ell_{\mathcal{B}} \rangle = \mathcal{A}[\mathbf{w}_{\hat{n}}]^\intercal \cdot \mathcal{B}[\mathbf{w}_{\hat{n}}]$$

*Proof.* Direct computation. □

**Lemma A2.** *Let $\text{vec}(\mathbf{w}_{\hat{n}})$ denote the $(D_1 D_2 \cdots D_{n-1} D_{n+1} \cdots D_n)$-vector given by*

$$\text{vec}(\mathbf{w}_{\hat{n}}) = \text{vec}(\mathbf{w}_1 \otimes \cdots \otimes \mathbf{w}_{n-1} \otimes \mathbf{w}_{n+1} \otimes \cdots \otimes \mathbf{w}_N)$$

*and suppose that the matricization has SVD $\mathbf{A}_{(n)} = \mathbf{U}^n \mathbf{\Sigma}^n (\mathbf{V}^n)^\intercal$. Then $\mathcal{A}[\mathbf{w}_{\hat{n}}]$ is the $D_n$-vector*

$$\mathcal{A}[\mathbf{w}_{\hat{n}}] = \mathbf{U}^n \mathbf{\Sigma}^n (\mathbf{V}^n)^\intercal \text{vec}(\mathbf{w}_{\hat{n}}).$$

*Proof.* Direct computation. □

**Lemma A3.**

$$\langle \boldsymbol{\pi}_n \nabla \ell_{\mathcal{A}}, \nabla \ell_{\mathcal{B}} \rangle$$
$$= \text{vec}(\mathbf{w}_{\hat{n}})^\intercal \mathbf{V}_{\mathcal{A}}^{(n)} \boldsymbol{\Sigma}_{\mathcal{A}}^n \boldsymbol{\Sigma}_{\mathcal{B}}^n (\mathbf{V}_{\mathcal{B}}^n)^\intercal \text{vec}(\mathbf{w}_{\hat{n}}).$$

*Proof.* By the above working,

$$\langle \boldsymbol{\pi}_n \nabla \ell_{\mathcal{A}}, \nabla \ell_{\mathcal{B}} \rangle$$
$$= \text{vec}(\mathbf{w}_{\hat{n}})^\intercal \mathbf{V}_{\mathcal{A}}^{(n)} \boldsymbol{\Sigma}_{\mathcal{A}}^n (\mathbf{U}_{\mathcal{A}}^n)^\intercal \mathbf{U}_{\mathcal{B}}^n \boldsymbol{\Sigma}_{\mathcal{B}}^n (\mathbf{V}_{\mathcal{B}}^n)^\intercal \text{vec}(\mathbf{w}_{\hat{n}}).$$

Recall that

$$(\mathbf{V}^n)^\intercal := \tilde{\mathbf{S}}_{(n)} \left( \mathbf{U}^{n+1} \otimes \cdots \mathbf{U}^N \otimes \mathbf{U}^1 \otimes \cdots \mathbf{U}^{n-1} \right)^\intercal$$

so that

$$\boldsymbol{\Sigma}^n (\mathbf{V}^n)^\intercal = \mathbf{S}_{(n)} \left( \mathbf{U}^{n+1} \otimes \cdots \mathbf{U}^N \otimes \mathbf{U}^1 \otimes \cdots \mathbf{U}^{n-1} \right)^\intercal$$

which we shorten to

$$\boldsymbol{\Sigma}^n (\mathbf{V}^n)^\intercal = \mathbf{S}_{(n)} \left( \tilde{\mathbf{U}}^{\hat{n}} \right)^\intercal$$

It follows that $\langle \boldsymbol{\pi}_n \nabla \ell_{\mathcal{A}}, \nabla \ell_{\mathcal{B}} \rangle$ equals

$$\text{vec}(\mathbf{w}_{\hat{n}})^\intercal \tilde{\mathbf{U}}_{\mathcal{A}}^{\hat{n}} \mathbf{S}_{(n)}^\intercal \mathbf{T}_{(n)} (\mathbf{U}_{\mathcal{B}}^{\hat{n}})^\intercal \text{vec}(\mathbf{w}_{\hat{n}}).$$

and the result follows. □

We are finally in a position to explain why simultaneous HOSVD is insufficient to guarantee safety. That is, we are in a position to explain why

$$\mathcal{A} = \mathcal{S} \times_1 \mathbf{U}^1 \times_2 \mathbf{U}^2 \cdots \times_N \mathbf{U}^N$$
$$\mathcal{B} = \mathcal{T} \times_1 \mathbf{U}^1 \times_2 \mathbf{U}^2 \cdots \times_N \mathbf{U}^N$$

does not guarantee $\langle \boldsymbol{\pi}_1(\nabla \ell_{\mathcal{A}}), \nabla \ell_{\mathcal{B}} \rangle \geq 0$. Our strategy for guaranteeing safety hinges on simultaneous diagonalization. However, HOSVD does not guarantee that the core $\mathcal{S}$ is diagonal. Instead, it introduces the condition of simultaneous orthogonality on subtensors of $\mathcal{S}$.

Unfortunately, there is nothing to guarantee that the cores $\mathcal{S}$ and $\mathcal{T}$ of two different tensors are "simultaneously all-orthogonal" across the two tensors. In other words, there is nothing to guarantee that the product of their matricizations

$$\mathbf{S}_{(n)}^\intercal \mathbf{T}_{(n)}$$

is diagonal in general. A specific setting where this holds is when the two tensors $\mathcal{A}$ and $\mathcal{B}$ admits a simultaneous tensor-SVD. It is an open question whether the condition can be guaranteed in a naturally-occurring more general setting.

## A6. Kickback is safe

Kickback is a complementary algorithm to feedback alignment that is motivated by the observation that neurons communicate via a single kind of signal, spikes, rather than using the two kinds of signals (the feedforward sweep and backpropagated errors) required by backprop (Balduzzi et al., 2015).

Kickback is a truncated version of backprop that computes gradient-estimates using the feedforward sweep together with global error/reward signals. The error signals take the form of a single scalar value broadcast to the entire network, for example via neuromodulators. One of the main results of the paper is that

**Theorem.** *If neurons are coherent then kickback is safe.*

*Proof.* See theorem 4 of Balduzzi et al. (2015). □

Coherence is, essentially, a positivity condition on synaptic weights that ensures the gradient-estimates computed by kickback have positive inner product with the gradients computed by backprop.

## A7. Comparison with Strongly-Typed RNNs

Typed linear algebra was proposed in Balduzzi & Ghifary (2016) (STNN), which applied the framework to analyze and simplify recurrent neural networks. The definition of typed vector space in this paper is more general than in STNN – it replaces an orthogonal basis with orthogonal projections.

No formal definition of strong-typing was provided in STNN. Informally, STNN stated: "We refer to architectures as strongly-typed when they both (i) preserve the type structure of their features and (ii) separate learned parameters from state-dependence". The definition of strong-typing in the main text, definition 5 is different from strong-typing in STNN, although it draws on similar intuitions. The settings are sufficiently disparate that no confusion should result.